\documentclass{amsart} %

\usepackage[usenames,dvipsnames,condensed]{xcolor}

\usepackage{amsthm}
\usepackage{amsfonts}
\usepackage[english]{babel}
\usepackage[usenames]{xcolor}
\usepackage{graphicx}
\usepackage{soul}
\usepackage{stfloats}
\usepackage{morefloats}
\usepackage{lscape}
\usepackage{epstopdf}
\usepackage{braket}
\usepackage{mathbbol}
\usepackage{tikz-cd}
\usepackage{shuffle}
\usepackage{booktabs}

\usepackage{algorithm,algorithmicx,algpseudocode}

\input{epsf.sty}
\usepackage{amsmath,amstext,amsopn,amsfonts,eucal,amssymb}
\input{epsf}
\usepackage{graphicx,wrapfig,url}

\usepackage[backend=bibtex,style=numeric]{biblatex}
\bibliography{refs} 

    \author[Zappala]{Emanuele Zappala$^{\ast}$} 
	\address{Yale University, New Haven, CT, USA} 
	\email{emanuele.zappala@yale.edu}

	\author[Fonseca]{Antonio Henrique de Oliveira Fonseca$^{\ast}$}
	\address[Co-first author]{Yale University, New Haven, CT, USA} 
	\email{antonio.fonseca@yale.edu}
	\thanks{$^\ast$Co-first authors.}
	
	\author[Moberly]{Andrew Henry Moberly}
	\address{Yale University, New Haven, CT, USA}
	\email{andrew.moberly@yale.edu}
	
	\author[Higley]{Michael James Higley}
	\address{Yale University, New Haven, CT, USA}
	\email{m.higley@yale.edu}
	
	\author[Abdallah]{Chadi Abdallah}
	\address{Baylor College of Medicine, Houston, TX, USA}
	\email{chadi.abdallah@bcm.edu}
	
	\author[Cardin]{Jessica Cardin}
	\address{Yale University, New Haven, CT, USA}
    \email{jess.cardin@yale.edu}
    
	\author[van Dijk]{David van Dijk} 
	\address{Yale University, New Haven, CT, USA} 
	\email{david.vandijk@yale.edu}

\begin{document}
	\title{Neural Integro-Differential Equations}

	\maketitle
	
\begin{abstract}
Modeling continuous dynamical systems from discretely sampled observations is a fundamental problem in data science. Often, such dynamics are the result of non-local processes that present an integral over time. As such, these systems are modeled with Integro-Differential Equations (IDEs); generalizations of differential equations that comprise both an integral and a differential component. For example, brain dynamics are not accurately modeled by differential equations since their behavior is non-Markovian, i.e. dynamics are in part dictated by history. Here, we introduce the Neural IDE (NIDE), a novel deep learning framework based on the theory of IDEs where integral operators are learned using neural networks.  
We test NIDE on several toy and brain activity datasets and demonstrate that NIDE outperforms other models. These tasks include time extrapolation as well as predicting dynamics from unseen initial conditions, which we test on whole-cortex activity recordings in freely behaving mice. Further, we show that NIDE can decompose dynamics into their Markovian and non-Markovian constituents via the learned integral operator, which we test on fMRI brain activity recordings of people on ketamine. Finally, the integrand of the integral operator provides a latent space that gives insight into the underlying dynamics, which we demonstrate on wide-field brain imaging recordings. Altogether, NIDE is a novel approach that enables modeling of complex non-local dynamics with neural networks.
\end{abstract}

	\section{Introduction}

Integro-differential equations (IDEs) are a class of functional differential equations that are non-local in time. IDEs naturally describe many dynamical systems, including population dynamics, nuclear reactor physics, and visco-elastic fluids, as considered in detail in \cite{Lak}. Further examples are models of brain dynamics \cite{Neural_fields,WC,amari1977dynamics}, as well as infectious disease spreading \cite{MK}.

Functional differential equations, and IDEs in particular, are equations determined by non-local operators. These are mappings, or {\it functionals}, from a space of functions into itself that require knowledge of the input function at non-infinitesimal neighbourhoods in order to be computed. Integral operators epitomize such functionals, and the corresponding theory of IDEs has become central, for instance, in kinetic theory \cite{GIKM}. Examples of such applications are the Boltzmann kinetic equation, the Vlasov equation, and the Landau kinetic equation. In contrast, differential operators such as time derivatives are local operators in that the notion of derivative at one point requires information from the ``immediate vicinity'' of the point. This is formalized through the concept of limit.

\begin{figure*}[t]
	\begin{center}
		\includegraphics[width=5.5in]{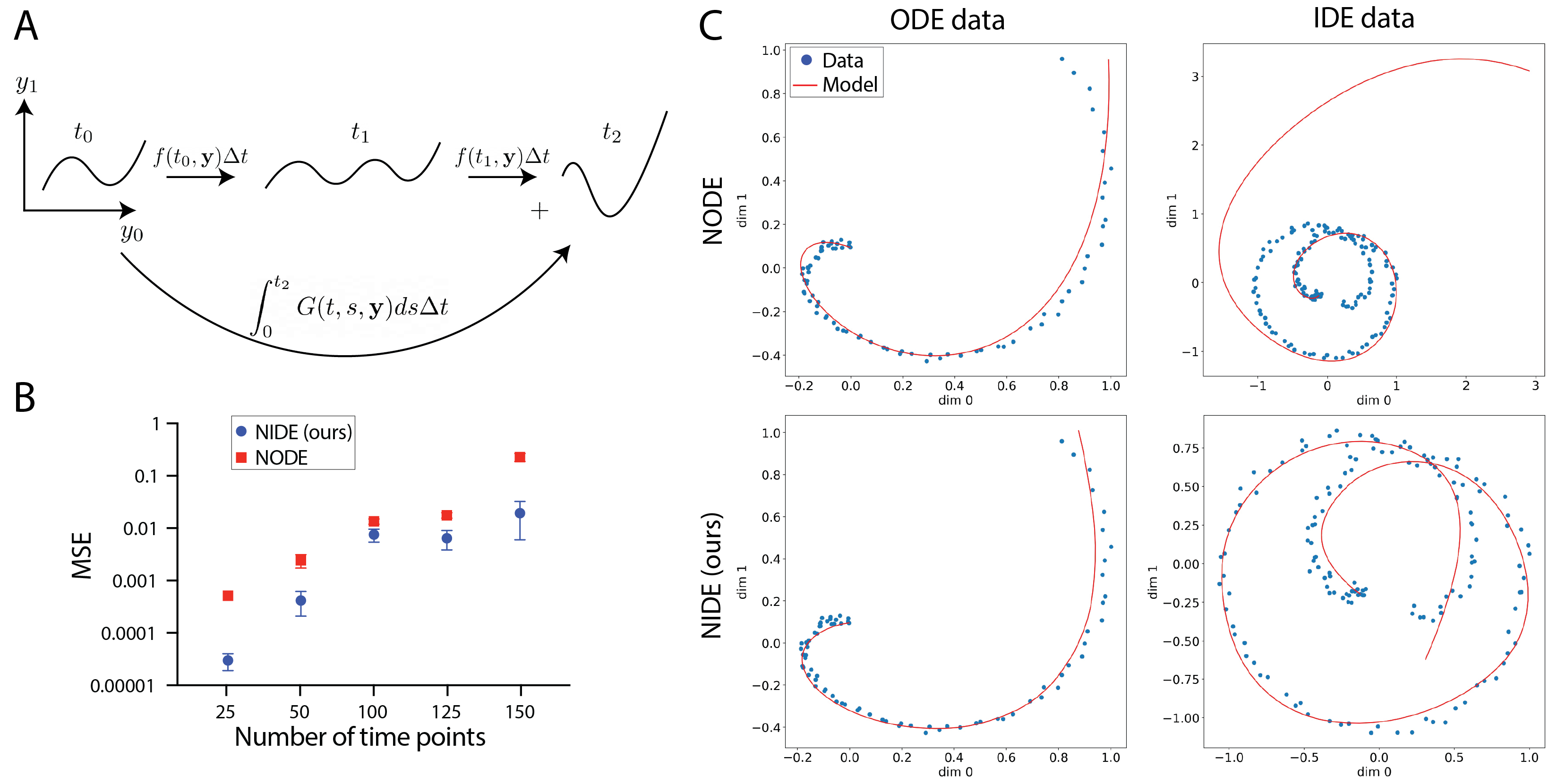}
	\end{center}
	\caption{Panel A shows a schematic of the NIDE method, where time steps are determined through a superposition of local and non-local components. Panel B shows performance (MSE of model fits) of NIDE and NODE on data sampled at increasingly longer time segments from an IDE system. NIDE significantly ($P<10^{-6}$) outperforms NODE, especially at longer time segments (see Table \ref{tab:spirals}). Panel C shows 2D spirals generated with either an ODE (left) or an IDE (right) and fitted by NODE (top) or NIDE (bottom). While both models can fit the ODE spiral, only NIDE can accurately fit the IDE spiral.}
	\label{fig:IDE_ODE_spirals}
\end{figure*}

The theory of IDEs stems from the necessity of modeling systems that present spatio-temporal relations which transcend local modeling, and works of Volterra on population dynamics have motivated this since as early as the beginning of the 1900's \cite{Volterra1,Volterra2}.
Consequently, IDEs present several properties that are unique to their purely non-local behavior \cite{GIKM,Enc_Math}. Despite their importance, no approach for learning IDE systems from data exists. Motivated by this, we develop a deep learning method called {\it Neural Integro-Differential Equation} (NIDE), which learns an IDE whose solution approximates data sampled from given non-local dynamics. A schematic representation of the functioning of NIDE is given in Figure~\ref{fig:IDE_ODE_spirals}A. To the best of our knowledge, this is the first deep learning framework for modeling of non-local continuous dynamics. 

\subsection{Contributions}
The main contributions of this article are as follows:
\begin{itemize}
	\setlength\itemsep{0.3em}
	\item We implement an IDE solver fully supported by PyTorch.
	\item We introduce a deep learning method for modeling IDEs, namely {\it Neural IDE}, which learns an integral operator from a space of functions into itself.
	\item We derive the adjoint state of NIDEs and use this for backpropagation during training.
	\item We present a method for decomposing dynamics into local and non-local components.
	\item Finally, we use our method to model brain dynamics, including wide-field calcium imaging in mice and fMRI recordings of people on ketamine.
\end{itemize}

\subsection{Related work}

The use of differential equation solvers to learn dynamics through neural networks has been pioneered in \cite{NODE,NODE2} to address the need of having continuous deep learning models. These models, called {\it Neural Ordinary Differential Equations}, will be referred to as NODEs in the rest of the article. 
Our theory introduces a deep learning approach to the case of functional differential equations, where dynamics present gloabl (i.e. non-local) behavior. An introduction to functional differential equations can be found in \cite{stech1981integral}. A viable approach to incorporating non-local dependence in ODEs would be, for example, the use of LSTMs in NODEs. However, this is essentially equivalent to the use of delay differential equations (DDEs \cite{DDEs}).

In various applications that span from computational biology to physics and engineering, however, both ODEs and DDEs are not capable of modeling dynamical systems with non-local properties, where the use of functional differential equations, such as IDEs, is employed instead \cite{Volterra1,Volterra2,GIKM,Waz^2,stech1981integral}.
In our setting, we learn an integral operator on a space of functions.  Learning operators on function spaces is a deep learning problem that has been considered for instance in \cite{DeepOnet} and \cite{DeepOnet_variation}, where the case of integral operators is considered in detail as well. The main difference between our method and the latter is that the use of an IDE solver, as we will show, allows us to learn in a continuous manner, as well as produce continuous dynamics.


Operator learning is a widely developed approach that encompasses numerous machine learning techniques, and is applied in several disciplines. We emphasize that the usual setting of operator learning is over a fixed grid that approximates the domain of the functions, and that the infinite dimensional space of functions is usually projected (e.g. Galerkin method). Moreover, recovering the continuous limit (i.e. the grid steps going to zero) is a fundamental challenge. A general perspective is given in \cite{Neural_Operator}, for instance, where the authors consider several problems arising from Partial Differential Equations (PDEs). In the introduction of \cite{Neural_Operator}, a relatively comprehensive account of the different methodologies is also provided. 

Solving IDEs by means of deep learning approaches has been considered in \cite{PIDEs}, and several methodologies are overviewed in \cite{DEEPXDE}. We point out, however, that the perspective of the present article is somehow inverted with respect to the previous works on IDEs appearing in the context of machine learning. In fact, to our knowledge, previous works have only focused on obtaining methods for solving IDEs, in supervised or unsupervised settings, that are given as input of the problem. Instead, the approach that we take learns an IDE whose solution approximates data sampled from a dynamical system. What is learned here is not a solution to a known system, but the system itself. Consequently, the input of our method is not an analytical IDE, but data without any a priori knowledge of the system that has generated it. The main novelty of the present work resides in the fact that no prior analytical knowledge of the system is required and,therefore, NIDE provides insight into dynamical systems whose underlying nature is not necessarily well understood.


\section{Learning dynamics with integro-differential equations}\label{sec:main}

We consider the problem of learning dynamics that depend nontrivially on non-local effects, as well as on instantaneous information. Such a dynamical system is expressed in terms of both the derivative of a function, and a temporal integral. The resulting equation of this type is an integro-differential equation, and its general form (see \cite{Lak,Waz^2}) is given as
$$
\frac{d\mathbf{y}}{dt} = f(t, \mathbf{y}) + \int_{\alpha(t)}^{\beta(t)} G(t,s,\mathbf{y}) ds,
$$

where $\mathbf{y} : \mathbb R \longrightarrow \mathbb R^n$ is a vector function of the independent time variable $t$. The functions $f : \mathbb R^{n+1} \longrightarrow \mathbb R^{n}$ and $G: \mathbb R^{n+2} \longrightarrow \mathbb R^n$ generate the dynamics. Observe that $G$ depends explicitly on $t$, so that differentiating the integral with respect to $t$ produces, in general, a new expression where an integral appears, using the Leibniz integral rule. The derivative is parameterized by a neural network $f$ along with a temporal integral of another neural network $G$, the main difference lying in the fact that past and future states explicitly appear in the present state. We call such a model {\it Neural Integro-Differential Equation}, or NIDE for short. The theory of IDEs is better understood in the setting where the function $G$ is a product of type $K(t,s)F(\mathbf y(s))$ for some arbitrary function $F : \mathbb R^n \longrightarrow \mathbb R^n$, and a matrix valued function $K : \mathbb R^2 \longrightarrow {\rm M}(\mathbb R,n)$, where $ {\rm M}(\mathbb R,n)$ indicates the space of square matrices with real coefficients. The function $K$ is called {\it kernel}.
In this article, we therefore consider systems of type
\begin{eqnarray}
	\frac{d\mathbf{y}}{dt} = f(t, \mathbf{y}) + \int_{\alpha(t)}^{\beta(t)}K(t,s) F(\mathbf{y}) ds, \label{eqn:NIDE_w_Kernel}
\end{eqnarray}

where $K$ and $F$ are both neural networks that will be learned during training, and $K$ will explicitly depend on the two time variables $t$ and $s$. In fact, we can generalize this setup and assume that the kernel function takes values in the space of rectangular matrices ${\rm M}(\mathbb R,m,n)$ where $m$ is the dimension of a latent space, and therefore  $F : \mathbb R^n \longrightarrow \mathbb R^m$ maps $\mathbf y$ in the latent space. This allows us to learn dynamics in the latent space. The kernel then maps the  latent space back to the original data space. We observe that the variables $t$ and $s$ that $K$ depends on can be intuitively interpreted as ``global'' and ``local'' times, respectively. or each time $t$, the value of the solution $\mathbf y(t)$ is determined via an integration in $ds$, so that for each value of $t$ we have a different integration space $[\alpha(t),\beta(t)]$ where $s$ lies.

We implement an IDE solver based on the work by \cite{GJSolver} and \cite{Karpel}. We use this solver  to obtain dynamics, i.e. a solution $\mathbf y(t)$ of an IDE as above, for given neural networks $K$ and $F$. To minimize the loss function $L$ with respect to target data, we use the adjoint state of the IDE in Equation~\ref{eqn:NIDE_w_Kernel}, cf. \cite{NODE,SFH}, or backpropagate directly through the IDE solver. The adjoint function is defined as $\mathbf a(t) := \frac{\partial L}{\partial \mathbf y}$ and its dynamics is determined by another system of IDEs, since it is obtained by differentiating the loss function evaluated on the output of the IDE solver. Before considering the dynamical system that determines the evolution of $\mathbf a(t)$, it is important to consider the method for solving the IDEs that is employed in this article, as this plays a fundamental role in the implementation and provides a substantial difference with respect to the case of NODEs. 

Notice that the functions $\alpha(t)$ and $\beta(t)$ are arbitrary and, therefore, for each $t \in [t_0,t_1]$, the derivative of the function $\mathbf y(t)$, $\frac{d\mathbf y}{dt}$, depends on values of $\mathbf y(t)$ both in the past, the present, and the future of the dynamics. Common choices of $\alpha$ and $\beta$ are $\alpha(t) = 0$ and $\beta(t)=t$ (Volterra IDEs) or $\alpha(t)=a$ and $\beta(t)=b$ (Fredholm IDEs), see Appendix~\ref{sec:IDE}.
Effectively, this means that the RHS of Equation~\ref{eqn:NIDE_w_Kernel} is a functional of $\mathbf y$ rather than a function. 
Therefore, an IDE solver cannot sequentially output the value of $\mathbf y(t)$ based on the computation of $\mathbf y$ on the previous time point. The idea is therefore to iteratively solve the IDE by producing successive approximations of the solution $\mathbf y(t)$ from an initial guess $\mathbf y_0(t)$ until convergence of $y_n(t)$ to the solution, within some predetermined tolerance. This allows us to have a full function $\mathbf y_n(t)$ at each iteration, and this can then be integrated over arbitrary intervals $[\alpha(t),\beta(t)]$ during the dynamics. 

Following the discussion in the previous paragraph, $\mathbf a(t) = \frac{\partial L}{\partial \mathbf y}$ is the functional derivative of the loss $L$ with respect to $\mathbf y(t)$. The loss function $L$ is computed by applying a chosen method (e.g. mean squared error) to the output of the IDE solver. One has the formula 
\begin{eqnarray}
\begin{aligned}
	\frac{d \mathbf a(t)}{dt} = -\int_{t_1}^{t_0}\mathbf a(t)^{T}\frac{\partial f}{\partial \mathbf y}dt \\
	- \int_{t_1}^{t_0}\mathbf a(t)^TK(t,t)F(\mathbf y(t))dt,
	\end{aligned}
\end{eqnarray}
which is derived in the next section. In order to derive the gradients with respect to the parameters $\theta$ of the neural networks $K$ and $F$ that define the NIDE we have another equation, namely 
\begin{eqnarray}
\begin{aligned}
	\frac{dL}{d\theta} = - \int_{t_1}^{t_0}\int_{\alpha(t)}^{\beta(t)} a(t)^T \frac{\partial K(t,s)}{\partial \theta} F(\mathbf y(s)) dsdt
	- \int_{t_1}^{t_0}\int_{\alpha(t)}^{\beta(t)} a(t)^T K(t,s) \frac{\partial F(s)}{\partial \theta} dsdt.
\end{aligned}
\end{eqnarray}
As discussed in Appendix~\ref{sec:Implementation}, 
the adjoint state can be both solved as an ODE where the RHS is obtained via integration of gradients, or it can be solved as an IDE through an iterative procedure. 

\section{Adjoint dynamics}\label{sec:Adjoint}

We now consider the dynamics of Equation~\ref{eqn:NIDE_w_Kernel} and derive the corresponding adjoint state, cf. \cite{NODE} and \cite{SFH}. In particular, we want to show that the adjoint function $\mathbf a(t) := \frac{\partial L}{\partial \mathbf y}$,  where $L$ is the loss function used for training, when $f$ is trivial, satisfies the IDE 
\begin{eqnarray}
	\label{eqn:adjoint}
	\frac{d\mathbf a_{aug}}{dt} = [-\mathbf a^T\cdot \frac{\partial }{\partial \mathbf y}(K(t,t)F(\mathbf y(t)))| 
	 - \mathbf a_{\mathbf \theta}^T\cdot \int_{\alpha(t)}^{\beta(t)} \frac{\partial }{\partial \mathbf \theta}(K(t,s)F(\mathbf y(s)))ds],
\end{eqnarray}
where $\mathbf a_{aug}$ is the adjoint state of the augmented dynamics, which includes the gradients with respect to the parameters $\mathbf \theta$ of the neural networks $K$ and $F$, and the square brackets refer to the fact that the dimensions of $\mathbf a_{aug}$ are split in two groups of direct summands, which we separate by a vertical bar for clarity. The case when $f$ is nontrivial is obtained from Equation~\ref{eqn:adjoint} by appropriately adding the computation for the adjoint found in \cite{NODE}. This accounts to adding $-\mathbf a^T\cdot \frac{\partial f(\mathbf y,t)}{\partial \mathbf y}$ to the first term in the RHS of Equation~\ref{eqn:adjoint}, and concatenating $-\mathbf a^t_\theta \cdot f(\mathbf y,t)$ to the second term, where $\theta$ here refers to the parameters of the neural network $f$. For simplicity we explicitly consider the case with trivial $f$, since the general case is a combination of this and the results of \cite{NODE}.  

Before showing that Equation~\ref{eqn:adjoint} describes the dynamics of the adjoint state, we recall the notion of functional derivative, and some known results that can be found in standard quantum field theory textbooks, e.g. \cite{QFT}. An application of such techniques in Bayesian inference has also recently appeared in \cite{kim2021fast}.

A functional $S$ is a mapping from a space of functions, say $\mathcal F$, to the real (or complex) numbers $S: \mathcal F\longrightarrow \mathbb R$. 
The notion of variation of a functional is a well known concept from the theory of calculus of variations, and finds its roots in the classical formulation of mechanics, where the equations of motion are found by minimizing the action, which is a functional, whose variation is set to zero. Extending the notion of variation to that of derivative for a functional is not straightforward, and it is formally defined as a limit of test functions (i.e. a distribution). Let us now consider the case of scalar argument functions in $\mathcal F$, since the extension to vector functions follows from this. For a functional of type $S(y(x)) = \int G(y(t),d_t y(t), \ldots , d^k_t y(t))dt$, where $d^r_t$ is a shorthand for $\frac{d^r}{dt^r}$, the functional derivative is 
\begin{eqnarray}
	\frac{\delta S(y(t))}{\delta y(t)} = \int [\frac{\partial G}{\partial y(\tau)} - d_\tau \frac{\partial G}{\partial (d_\tau y(\tau))} + \cdots ]\frac{\delta y(\tau)}{\delta y(t)}d\tau
\end{eqnarray}
The distribution $\frac{\delta y(\tau)}{\delta y(t)}$ is the Dirac's Delta function $\delta(t-\tau)$, so that one finds 
\begin{eqnarray}
	\frac{\delta S(y(t))}{\delta y(t)} = \frac{\partial G}{\partial y(t)} - d_\tau \frac{\partial G}{\partial (d_t y(\tau))} + \cdots 
\end{eqnarray}

Observe that, in our case, the integral part of the NIDE is an example of the functional $S$ given above, parametrized by global and local times, and without higher order derivatives with respect to $\mathbf y(t)$. Moreover, the function $G$ appearing in $S$ is the composition of $F$ and kernel $K$, so the functional derivative of the integral part of the NIDE in Equation~\ref{eqn:NIDE_w_Kernel} becomes 
\begin{eqnarray}\label{eqn:functional_der}
	\frac{\delta}{\delta(\mathbf y)}[ \int^{\beta(t)}_{\alpha(t)}K(t,s)F(\mathbf y(s))ds] = \frac{\partial( K(t,t)F(\mathbf y(t)))}{\partial y(t)}
\end{eqnarray}

We now consider the adjoint dynamics. We proceed in a fashion similar to Section~B.1 in \cite{NODE}. Recall, from above, that $\mathbf a(t) := \frac{\partial L}{\partial \mathbf y}$, where $L$ is the loss function evaluated on the output of the IDE solver. Then, we consider the equality $\frac{dL}{d\mathbf y(t)} = \frac{dL}{d\mathbf y(t+\epsilon)}\frac{d\mathbf y(t+\epsilon) }{d \mathbf y(t)}$ (cf. with previous discussion on functional derivatives), which corresponds to $\mathbf a(t) = \mathbf a(t+\epsilon)^T\cdot  \frac{d\mathbf y(t+\epsilon) }{d \mathbf y(t)}$, using the definition of $\mathbf a(t)$. To derive $\frac{d\mathbf y(t+\epsilon) }{d \mathbf y(t)}$, let us first write 
\begin{eqnarray}\label{eqn:y_epsilon}
	\mathbf y(t+\epsilon) = \mathbf y(t) + \int_t^{t+\epsilon}\int_{\alpha(\tau)}^{\beta(\tau)}K(\tau,s)F(\mathbf y(s))dsd\tau,
\end{eqnarray}
which is obtained by integrating the IDE from $t$ to $t+\epsilon$. For small values of $\epsilon$, Equation~\ref{eqn:y_epsilon} becomes 
\begin{eqnarray}\label{eqn:y_infinitesimal_epsilon}
	\mathbf y(t+\epsilon) = \mathbf y(t) + \epsilon \cdot \int_{\alpha(t)}^{\beta(t)}K(t,s)F(\mathbf y(s))ds,
\end{eqnarray}
from which we obtain 
\begin{eqnarray}
	\frac{d\mathbf y(t+\epsilon) }{d \mathbf y(t)} &=&  \frac{\delta}{\delta \mathbf y(t)}[\mathbf y(t) 
	 + \epsilon\cdot \int_{\alpha(t)}^{\beta(t)}K(t,s)F(\mathbf y(s))ds]\notag \\
	&=& 1 + \epsilon\cdot \frac{\partial( K(t,t)F(\mathbf y(t)))}{\partial \mathbf y(t)},
\end{eqnarray}
where in the last equality we have used Equation~\ref{eqn:functional_der}. As a consequence, for small $\epsilon$, we can write the equality
\begin{eqnarray}\label{eqn:a_t}
	a(t) =  a(t+\epsilon) + \epsilon a(t+\epsilon)^T\cdot \frac{\partial( K(t,t)F(\mathbf y(t)))}{\partial \mathbf y(t)},
\end{eqnarray}
up to terms that vanish to zero with the same order of $\epsilon^2$. 

To complete, we now compute the derivative of $\mathbf a(t)$ w.r.t. time. We have
\begin{eqnarray}
	\frac{d\mathbf a}{dt} &=& \lim_{\epsilon\to 0^+} \frac{\mathbf a(t+\epsilon) - a(t)}{\epsilon}\\
	&=& \lim_{\epsilon\to 0^+} -\frac{1}{\epsilon}\mathbf a(t+\epsilon)^T[\epsilon\frac{\partial( K(t,t)F(\mathbf y(t)))}{\partial y(t)}]\\
	&=& -a(t)^T\cdot \frac{\partial( K(t,t)F(\mathbf y(t)))}{\partial y(t)}.\label{eqn:adj_functional}
\end{eqnarray}

In order to update the parameters of the neural networks of the NIDE, we need to consider the augmented system (cf. \cite{NODE} and \cite{SFH}). The augmented state $\mathbf a_{\rm aug}(t)$ is obtained by considering the augmented IDE, where $\mathbf y_{\rm aug} = [\mathbf y(t) | \mathbf \theta]$ is obtained by concatenating the parameters of $F$ and $K$ to $\mathbf y(t)$. The temporal derivative $\mathbf \theta$ is trivial, since the parameters are time independent, and the augmented IDE reads
\begin{eqnarray}
	\frac{d \mathbf y_{\rm aug}}{dt} = [\int_{\alpha(t)}^{\beta(t)} K(t,s)F(\mathbf y(s))ds|\mathbf 0].
\end{eqnarray}
When considering the augmented adjoint function $\mathbf a_{\rm aug}(t)$, the time derivative of the dimensions of $\mathbf a_{\rm aug}(t)$ corresponding to $\int_{\alpha(t)}^{\beta(t)} K(t,s)F(\mathbf y(s))ds$ are obtained as above. Then, given $\mathbf a_{\mathbf \theta}(t) := \frac{\partial L}{\partial \mathbf \theta}$ we need to obtain the dynamics associated to $\mathbf a_{\mathbf \theta}(t)$. We can proceed as in the case of Equation~\ref{eqn:adj_functional} with the difference that the term involving $\frac{\delta}{\delta \mathbf \theta}$ in this case simply gives 
\begin{eqnarray}
\begin{aligned}
	\frac{\delta}{\delta \mathbf \theta} \int_{\alpha(t)}^{\beta(t)} K(t,s) F(\mathbf y(s)) ds
	= \int_{\alpha(t)}^{\beta(t)} \frac{\partial }{\partial \mathbf \theta}(K(t,s)F(\mathbf y(s)))ds.
\end{aligned}
\end{eqnarray}
The latter then can be used to compute the time derivative $\mathbf a_{\theta}(t)$ completing the derivation of Equation~\ref{eqn:adjoint}.





\section{Experiments}

In this section we perform several experiments with NIDEs on data that was analytically generated by dynamical systems corresponding to IDEs, and data of real-world systems that present non-local characteristics. Specifically, we consider brain dynamics data from whole-cortex activity recordings in freely behaving mice, and fMRI brain activity recordings of people on ketamine.
We focus on comparing our model with NODEs, as this is another model that is continuous in time, and with LSTMs, which are discrete-time deep learning methods that model non-locality by including previous states of the data. 

In addition, we investigate the interpretability of NIDEs in two ways: 1) By decomposing the dynamics into instantaneous and non-instantaneous components, and 2) by inspecting the latent space provided by the integrand function of the integral operator $F$.

We analyze, in detail, the capability of NIDE to extrapolate with respect to time, and to generalize to unseen initial conditions. Furthermore, in order to directly compare the expressivity of NIDE  to NODE, we consider fitting models to data sampled from increasingly complex dynamics. We show that when the dynamics are generated by an IDE with sufficiently complex non-local properties, NODE is not capable of properly fitting the data, but NIDE is. Details about the architecture of the model used in each task are provided in Table \ref{tab:Hyperparameters}. 

\subsection{Testing the expressivity of NIDE}

We first compare the expressivity of NIDE to NODE by fitting data of dynamics generated by IDE and ODE systems. 
Here, we use the same number of parameters for both NIDE and NODE, with NIDE having its parameters split between function $F$ and kernel $K$. While both NIDE and NODE perform well on the ODE generated curve, only NIDE can accurately fit the IDE generated data (Figure~\ref{fig:IDE_ODE_spirals}C). These experiments suggest that, with the same number of parameters, NIDE is a more expressive model than NODE, capable of fitting complex non-local dynamics.

To further explore this, we perform a more systematic experiment in which we gradually increase the data complexity, while fitting both NIDE and NODE models. Specifically, we sample points from a self-intersecting spiral that was generated using an IDE and fit models while increasing the range from which we sample the dynamics. We observe that shorter, and thus simpler dynamics, can easily  be fitted by both models while longer, and thus more complex dynamics, can only be properly fitted by NIDE (Figure \ref{fig:IDE_ODE_spirals}B, Table \ref{tab:spirals}).

\begin{table*}[t]
	\caption{\ Average MSE for NIDE (ours) and NODE on fitting data sampled from an IDE generated self-intersecting spiral (lower is better).}
	\label{tab:spirals}
        \centering
        \resizebox{0.95\textwidth}{!}{
		\begin{tabular}{cccccc}
			\toprule
			& \multicolumn{5}{c}{Number of time points}                   \\
			\cmidrule(r){2-6}
			& 25 & 50 & 100 & 125 & 150 \\
			\midrule
			NODE & 5.0E-04$\pm$ 1.0E-04 & 2.4E{-03}$\pm$ 7.0E-04 & 1.35E-01$\pm$1.6E-03 & 1.7E-02$\pm$2.1E-03 & 2.34E-01$\pm$4.2E-02\\
			NIDE & \textbf{2.82E-05$\pm$ 1.01E-05} & \textbf{4.02E-04$\pm$ 2.00E-04} & \textbf{7.5E-03$\pm$ 2.10E-03} & \textbf{6.4E-03$\pm$2.60E-03} & \textbf{1.94E-02$\pm$1.34E-02}\\
			
		\end{tabular}
 		}
\end{table*}

\subsection{Time extrapolation}

\begin{figure*}[t]
	\begin{center}
		\includegraphics[width=1\columnwidth]{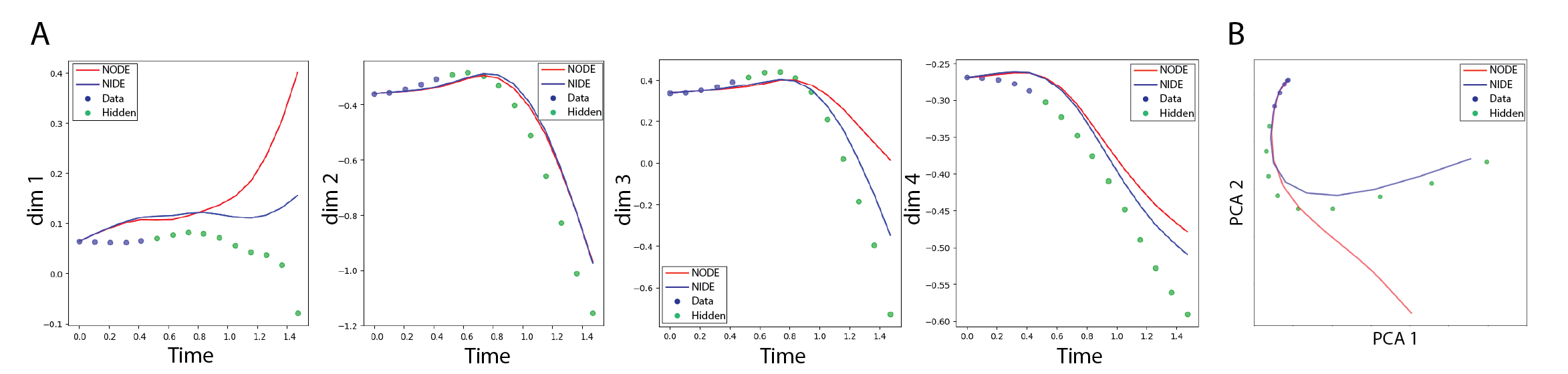}
	\end{center}
	\caption{Comparison between NODE (red) and NIDE (ours, blue) in a time extrapolation task. Shown is one of the $4D$ curves. Panel A shows each of the $4$ dimensions as a function of time, while panel B shows the 2D PCA projection of the dynamics. Data points that were masked during training and were then extrapolated are indicated as ``hidden'' (green). NIDE outperforms NODE in a time extrapolation task where, during training, up to $50\%$ of the points are masked and then predicted during inference.}
	\label{fig:extrapolation}
\end{figure*}

To assess the ability of NIDE to generalize to future time points (i.e. extrapolate), we generate 1000 $4D$ dynamics over a fixed interval using an IDE with uniformly sampled initial conditions. During training, we randomly mask up to $50\%$ of the time points at the end of the dynamics, which we predict during test. The experiments show that NIDE has lower MSE (Table \ref{tab:4d_curves_extrapolation}) and higher $R^2$ for the masked points, thereby demonstrating that NIDE extrapolates better than NODE on data sampled from an IDE system. An example of such extrapolation is shown in Figure~\ref{fig:extrapolation}.

Next, to inspect the extent to which NIDE can generalize to new initial conditions, we consider the model trained on the $4D$ curves dataset and evaluate it on curves from new initial conditions that have not been seen during training. We find that NIDE yields lower MSE for predicted dynamics from initial conditions than NODE, as shown in Table \ref{tab:4d_new_initial_conditions} per extrapolated time point. 

\subsection{Decomposition into Markovian and non-Markovian dynamics}

NIDEs learn dynamics through the derivative of the function $\mathbf y(t)$ considered above, as the superposition of two components. The first one is common to the setting of NODEs, where $\frac{d \mathbf y}{dt}$ is given by a neural network $f$, evaluated on pairs $(t,\mathbf y(t))$: $f(t,\mathbf y)$. We refer to this summand as the {\it Markovian} component, as the contribution of $f$ at the instant $t$ depends only on the current time step of the solver. On the contrary, the integral part of the NIDE will be referred to as the {\it non-Markovian} component, as at each time $t$ this depends on the full interval $[\alpha(t),\beta(t)]$. 


To test this construction, we generate $2D$ curves through an IDE with nontrivial analytical functions $f, K, F$, with different initial conditions. Since $f, K$ and $F$ are known, we can recover the decomposition into Markovian and non-Markovian for each of the curves in the dataset. This gives us a ground truth. 

We train a NIDE model on the full data without providing any information regarding the underlying decomposition. Then, we compute the Markovian and non-Markovian decomposition of the trained models, and compare these to the ground truth. We find that NIDE can accurately reconstruct the ground truth components for both training and validation data. We find $R^2 = 0.991\pm0.01$ (mean$\pm$std) for the whole fitting, $R^2 = 0.957\pm0.03$ for the instantaneous part, and $R^2 = 0.943\pm0.05$ for the integral part.
An example of a reconstructed decomposition is shown in Figure~\ref{fig:markovian_splitting}.


We stress that this is indeed a highly nontrivial result, as there is no a priori reason to assume that a given dynamics needs to be learned following a prescribed superposition of Markovian and non-Markovian components. However, if NIDE is exposed to a sufficiently large sample of dynamics generated by a given system, it is able to accurately reconstruct the local and non-local components.

We note that, for this experiment, we did not compare NIDE to other models since, to the best of our knowledge, no prior attempts have been made to decompose dynamics into their Markovian and non-Markovian components. For example, a NODE would only learn a Markovian component.

\subsection{Modeling brain dynamics using NIDEs}

Spatiotemporal neural activity patterns can be modeled as a dynamical system \cite{vyas2020computation}, for example using ODEs \cite{linden2021go}. However, ODEs have shown only partial success in modeling brain dynamics, arguably due to the significant non-local component of neuronal activity \cite{linden2021go,breakspear2017dynamic}. Previous attempts at considering the non-local behavior of brain dynamics include the Amari model \cite{amari1977dynamics}. The latter was a theoretical model based on IDEs, and NIDE provides a practical and concrete realization of it.
 More specifically, we apply NIDE to wide-field calcium brain imaging in freely behaving mice as well as fMRI recordings in humans on ketamine.

Wide-field calcium imaging is a recently developed technology that uses calcium indicators as a proxy for neural activity, which allows recording of brain dynamics at high spatial and temporal resolution in mice \cite{cardin2020mesoscopic}. Details about data collection and pre-processing can be found in Appendix~\ref{sec:exp_details} and Appendix~\ref{sec:data_collection}.

\begin{figure*}[t]
	\begin{center}
		\includegraphics[width=5.5in]{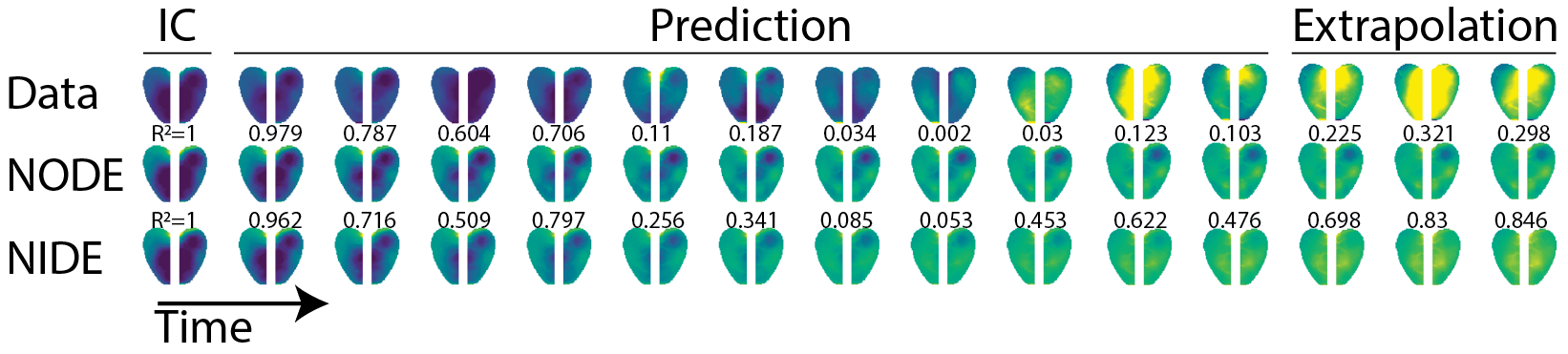}
	\end{center}
	\caption{Time extrapolation performance of NODE and NIDE (ours) on wide-field calcium brain imaging data. NODE and NIDE were trained on brain dynamics in which time points (at the end of sequences) were masked. Time extrapolation was then assessed by predicting the masked (future) time points. NIDE (ours) outperforms NODE in this task. Shown is a dynamics with NODE and NIDE predictions and corresponding $R^2$ values with the ground truth. IC refers to the initial condition, which is provided to both models.}
	\label{fig:brain_extrapolation}
\end{figure*}

We compare NIDE to NODE in a time extrapolation task of calcium imaging recordings in mice passively exposed to a visual stimulus \cite{lohani2020dual}. Extrapolation performance is shown as the mean $R^2$ between prediction and ground truth for several recording segments (Table \ref{tab:brain_extrapolation}). We find that NIDE outperforms NODE in this task. An example of an extrapolation is provided in Figure~\ref{fig:brain_extrapolation}. Qualitatively, we observe that NIDE learns a better model because it predicts excitation in the visual cortex, which is present in the original data, while NODE predicts an inhibitory pattern.

We also test the performance of NIDE on predicting dynamics of new initial conditions (see Figure~\ref{fig:brain_new_init}). In this task, the whole sequence is predicted from a given initial condition. Here, too, NIDE outperforms NODE. 


The function $F$ (the integrand of the integral operator) produces a latent space embedding of the data. To test the resulting latent representation we compare the output of $F$ to PCA and UMAP \cite{UMAP} on the calcium imaging data. As shown in Figure~\ref{fig:brain_latent}, the latent space of $F$ presents a higher degree of geometric structure and resembles more accurately the topology of the embedded manifold with respect to time, which we quantify with k-NN regression using k=3 neighbors (Figure \ref{fig:brain_latent}).

Next, we apply our method of Markovian and non-Markovian decomposition to the calcium imaging data. This allows us to quantify the extent to which the brain dynamics are dictated by non-local components (Figure~\ref{fig:brain_markovian}). Interestingly, the decomposition shows that the Markovian part of the signal mainly represents inhibition of neural activity in the visual cortex, while the non-Markovian part represents excitation, suggesting that the visual cortex is driven by non-local dynamics. Neural inhibition after the excitation is a known property of the excitatory-inhibitory networks in the brain, ensuring that any increase of neural activity will be followed by suppression \cite{bhatia2019precise}. Its Markovian dynamics suggests that excitation is primarily determined locally in time, thus by the current state of the brain.

Functional magnetic resonance imaging (fMRI) is the most common technique for studying whole-brain activity in humans. In fMRI, the blood-oxygen-level dependent (mathbf) signal is used as a proxy for brain activity \cite{akiki2019determining}. Major depression is a worldwide leading cause of disability with poorly understood neurobiology and high treatment resistance. Ketamine, a serendipitously discovered rapid acting antidepressant, offers a unique opportunity to unravel the brain networks underlying major depression and to establish novel treatment targets \cite{abdallah2018neurobiology}. In this experiment, fMRI scans were acquired repeatedly during infusion of normal saline followed by intravenous infusion of a subanesthetic dose of ketamine in healthy subjects. The aim is to determine the ketamine induced brain dynamics during infusion, which are believed to reverse critical aspects of the psychopathology of major depression and lead to sustained relief of symptoms \cite{abdallah2017ketamine,abdallah2021robust}. 

We find that the decomposition into Markovian and non-Markovian components, learned by NIDE, provides a space with better separation between ketamine and control conditions, as indicated by a higher k-NN classification accuracy with k=3 neighbors (Figure \ref{fig:fmri_latent}, Table \ref{tab:knn_performances}). This suggests that NIDE recovers dynamics relevant to the underlying biology. Interestingly, the separation between the two conditions is greater in the instantaneous part, compared to the integral part, suggesting that ketamine affects the Markovian component of brain dynamics, possibly by inducing a brain state that is more localized in time. 


\section{Scope and limitations}

The objective of this work is to introduce a new machine learning framework for IDEs. As such, this is applicable to several domains arising in biology, physics, and engineering where data is sampled from systems that follow IDE-like dynamics.

The model introduced in this article is based on temporal integrals. It is of great interest to extend the present theory to the study of Partial IDEs, i.e. IDEs where both spatial and temporal integrals appear. The iterative nature of the IDE solver considered here introduces a hyper-parameter, namely the number of iterations. While in our experiments using a low number of iterations sufficed, fine tuning the number of iterations might be a nontrivial task. NIDE uses an IDE solver which integrates locally at each time point. While our implementation of integration makes use of {\it torchquad} \cite{gomez2021torchquad} and it is therefore fully supported by GPUs, hence relatively efficient, improvements can be made by implementing more advanced and more efficient integration methods.

\section{Conclusions}
We presented a novel method for modeling dynamical systems with non-instantaneous behavior. Our method, termed Neural Integro-Differential Equations (NIDE), models complex non-local dynamics based on the theory of IDEs, which are commonplace in physics and biology. 
To train our model, we have presented a differentiable IDE solver implemented in PyTorch. We have performed extensive experimentation on tasks such as time extrapolation and predicting dynamics of unseen initial conditions on both toy and real world data and show that NIDE outperforms other methods. To showcase real-world applications, we have used NIDE to model two brain activity recording datasets: 1) Wide-field calcium brain imaging in freely behaving mice, and 2) fMRI recordings of people on ketamine. For the calcium imaging dataset, NIDE more accurately predicts future brain states as well as unseen dynamics. In addition, a latent space learned by NIDE, via the integrand of the integral operator, provides an embedding that more accurately reflects time dynamics compared to other unsupervised embedding methods. For the fMRI data, NIDE learns a decomposition into instantaneous and non-instantaneous dynamics that improves the ability to distinguish between saline (control) and ketamine conditions suggesting that NIDE infers components of the dynamics that reflect the underlying biology. IDEs are a well established mathematical framework and have found numerous applications in mathematics, physics, and biology. For example, certain problems in physics, including plasma and nuclear reactor physics, are formulated by means of IDEs. In addition, infectious disease spreading as well as population dynamics are often modeled with IDEs. Here, we have taken IDEs into the world of deep learning, and since many real-world dynamical systems display non-instantaneous (non-Markovian) behavior, we anticipate that there will be many applications for NIDEs.

\section*{References}
\vspace{-1em}
\begin{footnotesize}
	\printbibliography[heading=none]
\end{footnotesize}

        
\appendix

\section{Integro-Differential Equations}
\label{sec:IDE}

Integro-differential equations (IDEs) are an important class of functional differential equations where an indeterminate function $\mathbf y(t)$ appears both under the sign of integral and as the argument of the differential operator $\frac{d}{dt}$. IDEs have found deep applications in computational biology, physics, engineering and other applied sciences \cite{GIKM,SG,CS,MK}. Although the theory of integral and integro differential equations is strongly connected to the theory of differential equations, as well as delay differential equations, it is well known that they are not equivalent. Reference \cite{GIKM} provides a comprehensive perspective on the difference between local equations (ODEs) and non-local equations (IDEs), with a presentation of important group theoretic methodologies that have been adapted from the study of symmetries of space of solutions of differential equations, to the theory of IDEs. In addition, applications of IDEs to physics are discussed in detail. 

The typical expression of an IDE is given by 
\begin{eqnarray}\label{eqn:IDE}
	\frac{d^k\mathbf{y}}{dt^k} = f(t, \mathbf{y}) + \int_{\alpha(t)}^{\alpha(t)} G(t,s,\mathbf{y}) ds,
\end{eqnarray}

where $G(t,s,\mathbf{y}) $ is often assumed to split in the product $K(t,s)F(\mathbf y(s))$, where $K$, called the kernel, is a matrix valued function of $t$ and $s$, and $F$ is a (generally non-linear) function $F: \mathbb R^n \longrightarrow \mathbb R^n$. When $F$ is linear, the equation is said to be a linear IDE, while when $F$ is nonlinear, the IDE is said to be nonlinear. Two important families of IDEs are the Volterra and the Fredholm equations. Volterra equations correspond to the choice of $\alpha(t) = a$ for some $a$, and $\beta(t) = t$ for all $t$, while Fredholm equations are given by $\alpha(t) = a$ and $\beta(t) = b$ for all $a,b$. In other words, in a Volterra IDE, the interval of integration grows linearly during the dynamics, while in a Fredholm IDE the interval of integration is held fixed during the whole dynamics. In particular, for a Fredholm IDE the system depends on the past, as well as present and future states of the system. 

The existence and uniqueness problem for IDEs is, similarly to the case of ODEs and PDEs, one of the main questions in the theory. There exist numerous general results, obtained for example applying the Schauder or Tychonoff fixed point theorems, that guarantee that under suitable regularity conditions on the function $G$ above, the initial value problem for Equation~\ref{eqn:IDE} admits solutions. The solutions are also unique and continuous with respect to the initial data under additional hypotheses. We refer the reader to Chapter 1 of \cite{Lak} for the details regarding the statements and proofs of these fundamental results. In addition, several methods for solving IDEs are given in \cite{GIKM,Waz^2,Zem}.

\section{Implementation}\label{sec:Implementation}

Here we discuss the implementation of the method, with particular emphasis on the adjoint state, and the computation of the terms appearing in Equation~\ref{eqn:adjoint}. Observe that, following the formulation of the adjoint state in \cite{NODE}, $\mathbf y(t)$ corresponds to the first $n$ dimensions of the adjoint $\mathbf a(t)$, where $n$ is the dimension of the dynamics. This was employed in \cite{NODE} in order to compute the dynamics again in reverse time during the backward pass. Therefore, Equation~\ref{eqn:adjoint} is an IDE, since the RHS depends on (some of the dimensions of) $\mathbf a(t)$ along the whole trajectory, and they are integrated. This equation, therefore, should in principle be solved as an IDE by the iteration procedure described in the first Appendix. 
However, this might require several iterations in order to accurately compute the gradients. In addition, it is not clear in principle how to set the number of iterations in order to reduce the complexity for the computation of the backward pass, and the iterations needed during the forward and backward in order to reach a determined precision might be different. 

To by-pass this potential bottleneck in the training of NIDEs, we save the  values of $\mathbf y$ computed in the forward pass at the data points, and utilize them to compute the partial derivatives $\frac{\partial }{\partial \mathbf y}(K(t,t)F(\mathbf y(t)))$ and $\frac{\partial }{\partial \theta}(K(t,s)F(\mathbf y(s)))$. In practice, this is done by applying the Jacobian-vector product, i.e. the function torch.autograd.grad where $\mathbf a$ is used for grad outputs.
We interpolate $\mathbf y$ between data points in order to evaluate $\mathbf y$ on arbitrary global and local times, i.e. $t$ and $s$ respectively. 

Observe that integrating torch.autograd.grad with $\mathbf a$ as grad outputs, produces the inner product $\int^{\beta(t)}_{\alpha(t)}\mathbf a^T\cdot \frac{\partial }{\partial \mathbf \theta}(K(t,s)F(\mathbf y(s)))$, where $\mathbf y$ is interpolated for evaluation in $s$. Since $\mathbf a$ depends only on what we have indicated as global time $t$, and not on local time $s$, it follows that the integration takes the form reported in the RHS of Equation~\ref{eqn:adjoint}, since $\mathbf a$ is constant during the integration in $ds$. Integration is efficiently performed using the integrator torchquad, developed by the European Space Agency (ESA). This can be found at https://github.com/esa/torchquad. 

To implement the backward, i.e. the adjoint state as described in this article, we introduce an nn.Module called IDEF (for Integro-Differential Equation Function) which outputs the components needed to compute the adjoint state. These components are then used to compute the adjoint dynamics (cf. NODE) which is then solved using an ODE solver. 

Algorithm~\ref{Algo:Solver} shows the functioning of the IDE solver, while Algorithm~\ref{Algo:NIDE} summarizes NIDE, the method introduced in this article. 

\begin{algorithm}
    \caption{IDE solver based on Adomian decomposition}
    \label{Algo:Solver}
    \begin{algorithmic}[1]
        \Require{$\mathbf y_0$} \Comment{Initial condition}
        \Ensure{$\mathbf y(t)$} \Comment{Solution to IDE with initial $\mathbf y_0$}
        \State{$\mathbf y^0(t) := \mathbf y_0$} \Comment{Initial solution guess}
        \While{${\rm iter}\leq {\rm max ite}r$ and ${\rm error} > {\rm tolerance}$}
        \State{Solve ode: $\frac{d\mathbf y}{dt} = f(\mathbf y^i,t) + \int_{\alpha(t)}^{\beta(t)}K(t,s)F(\mathbf y^i(s)ds$}
        \State{Set solution to be: $\mathbf y^{i+1}$}
        \EndWhile
    \end{algorithmic}  
\end{algorithm}
 
  \begin{algorithm*}
  \caption{Neural IDE training step}
  \label{Algo:NIDE}
  \begin{algorithmic}
  \Require{$K, F, f$ and $X$} \Comment{Neural networks and data} 
  \Ensure{$K, F, f$}\Comment{Dynamical IDE system trained over data $X$}
  \State{$\mathbf y_0$}\Comment{Initial condition from $X$}
  \State{Solve IDE using Algorithm~\ref{Algo:Solver}}
  \State{$L(\mathbf y(t),X)$}\Comment{Compute loss}
  \State{$\mathbf y_{\rm aug} := [\mathbf y|\theta]$}\Comment{Concatenate parameters $\theta$}
  \State{$\mathbf a:= \frac{\partial L}{\partial \mathbf y}$, $\mathbf a_\theta=\frac{\partial L}{\partial \theta}$}
  \State{$\frac{\partial }{\partial \mathbf y}(K(t,t)F(\mathbf y(t)))$ and $\frac{\partial }{\partial \theta}(K(t,s)F(\mathbf y(s)))$}\Comment{Computed using Autograd} 
  \State{$\int_{\alpha(t)}^{\beta(t)}\frac{\partial }{\partial \theta}(K(t,s)F(\mathbf y(s)))ds$}\Comment{Use interpolation and integrator}
  \State{Solve Eqn~\ref{eqn:adjoint}}\Comment{Use previous steps to recast Eqn~\ref{eqn:adjoint} as an ODE}
  \end{algorithmic}
  \end{algorithm*}

\section{Additional details on the experiments and hyperparameters}\label{sec:exp_details}

The hyperparameters for the experiments were as follows. The initial learning rate (lr) was $10^{-3}$ and we have used a ``cosine annealing'' scheduler that produced oscillations of the lr between $10^{-3}$ and $10^{-7}$ with period of $50$ epochs. The data dynamics was downsampled to $20$ points for each dynamics and the integrator of the integral operator sampled (through Monte-Carlo sampling) $1000$ points in the integration domain for each time step.  The Adam optimizer was used in all experiments. The architectures of the analytical splitting into Markovian and non-Markovian components consisted of two layers networks with a total of $344$ parameters ($182$ for the kernel and $162$ for the $F$ function). All the other details were the same as described above. The architectures of the neural networks used are summarized in Table~\ref{tab:Hyperparameters}. In the table, free function refers to the non-instantaneous part of the system, while $F$ and kernel refer to the components of the integral operators. 

\begin{figure*}[htb]
	\begin{center}
		\includegraphics[width=5in]{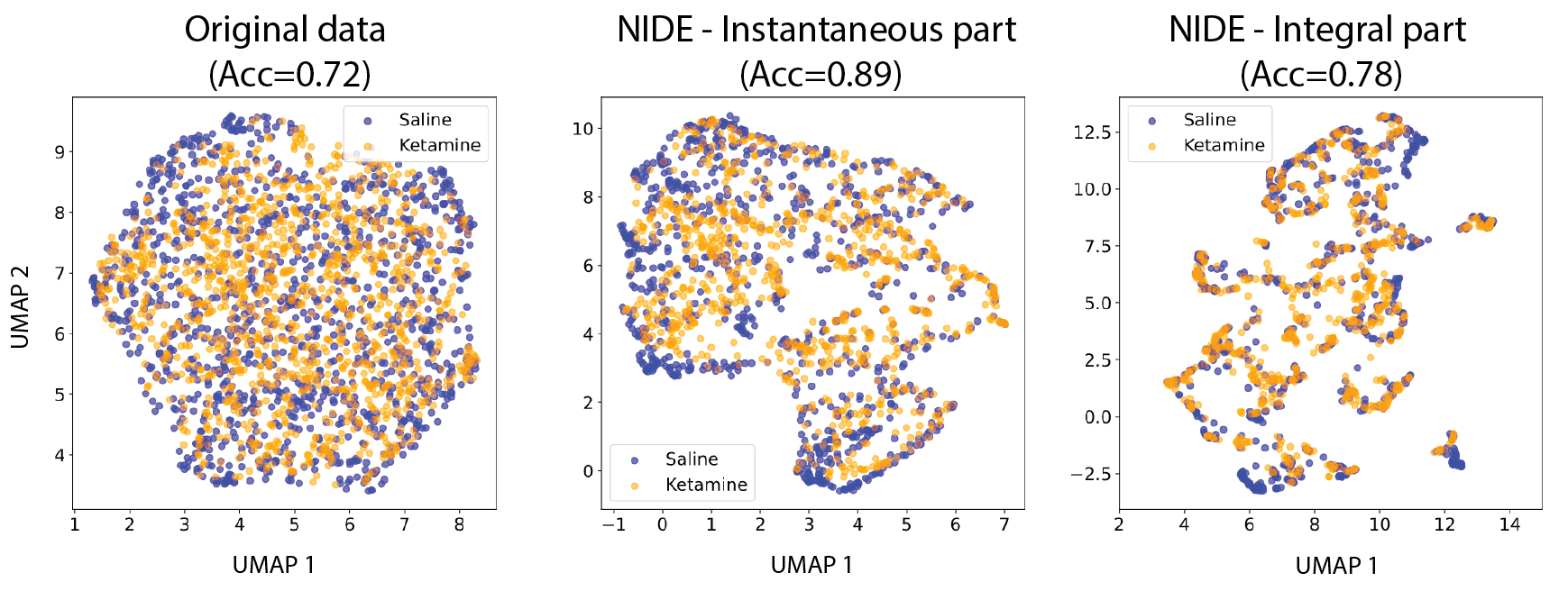}
	\end{center}
	\caption{fMRI brain recordings of people on ketamine (orange) or saline (blue, control). Each point is one time point. Shown is a comparison between the UMAP embeddings of the fMRI recordings of the original data (left), the instantaneous part of NIDE (middle), and the integral part of NIDE (right). Compared to the raw data, NIDE reveals more structure as well as increases the separation between the two conditions, in particular in the instantaneous part, as quantified using k-NN classification accuracy using k=3 (performance shown under the panel titles). The performance for other number of neighbors is shown in Table \ref{tab:knn_performances}}
	\label{fig:fmri_latent}
\end{figure*}

The fMRI data consists of 424 voxels recorded for 1240 time points for a subject under a Saline injection and 1240 time points under Ketamine, which therefore determines a $424D$ original data space. The k-nearest neighbors algorithm (k-NN) for the whole dynamics is applied in this space. For training, we reduce the dimensionality of the fMRI data to its first 10 principal components, which represents $\approx 50\%$ of the data's explained variance. Also, the whole recording is divided in segments of 15 time points, thus forming a total of 82 segments (dynamics). 70\% of the segments are used for training while the remaining 30\% are held out for validation. Instantaneous and integral dynamics latent dimensions are therefore 10D and k-NN for these components is computed in this space. In all these cases, 2D UMAP is used as means for data visualization. These three different terms correspond (in the same order to the three panels in Figure~\ref{fig:fmri_latent}. Further details about the fMRI data used in our experiments can be found in \cite{nemati2020unique}.

The calcium imaging recording consists of 3k frames with resolution of 256 $\times$ 256 pixels, which is reduced to $10D$ via PCA as described in the Data Collection. The whole recording is divided in segments of 15 time points, thus forming a total of 200 segments (dynamics). 70\% of the segments are used for training while the remaining 30\% are held out for validation.

\section{Supplementary results}\label{sec:sup_results}
We report the results on the embedding of the map $F$ (integrand of the learned integral operator) on fMRI data, further results regarding time extrapolation and new initial condition of NIDE and NODE on calcium imaging data, and the Markovian/non-Markovian decomposition of NIDE on calcium imaging data. 

Regarding NIDE's performance on the calcium imaging dataset, we show in Table~\ref{tab:brain_extrapolation} the mean $R^2$ for extrapolated points of the dynamics. Here we represent only time points that were not seen by the models during training, while the initial part of the dynamics, used during training is not reported.  The table shows that due to the non-local nature of the underlying dynamics, NIDE presents lower extrapolation errors.

\begin{figure}[htb]
	\begin{center}
		\includegraphics[width=1\columnwidth]{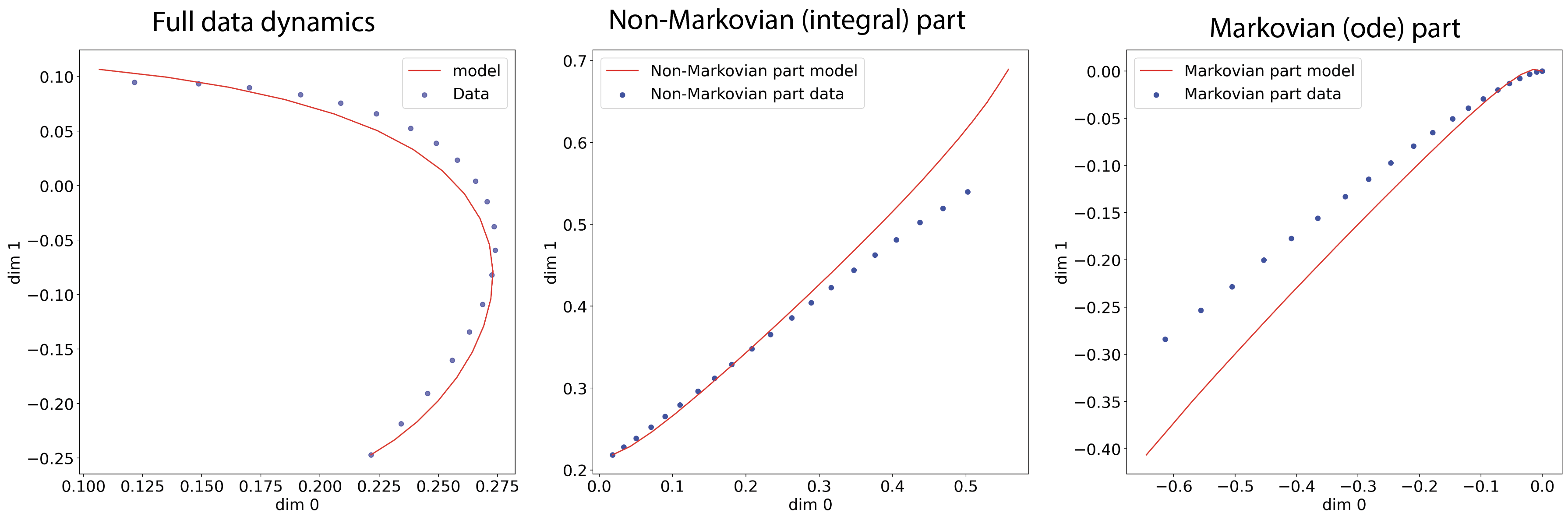}
	\end{center}
	\caption{Decomposition into Markovian and non-Markovian components of $2D$ dynamics using NIDE. Shown are dynamics generated using an IDE that has both ODE (Markovian) and integral (non-Markovian) parts. Both training data (points) and model fit (red line) are shown. NIDE fits the complete dynamics (left), and also recovers the ODE (middle) and integral (right) parts.}
	\label{fig:markovian_splitting}
\end{figure}

\begin{figure*}[htb]
	\begin{center}
		\includegraphics[width=5in]{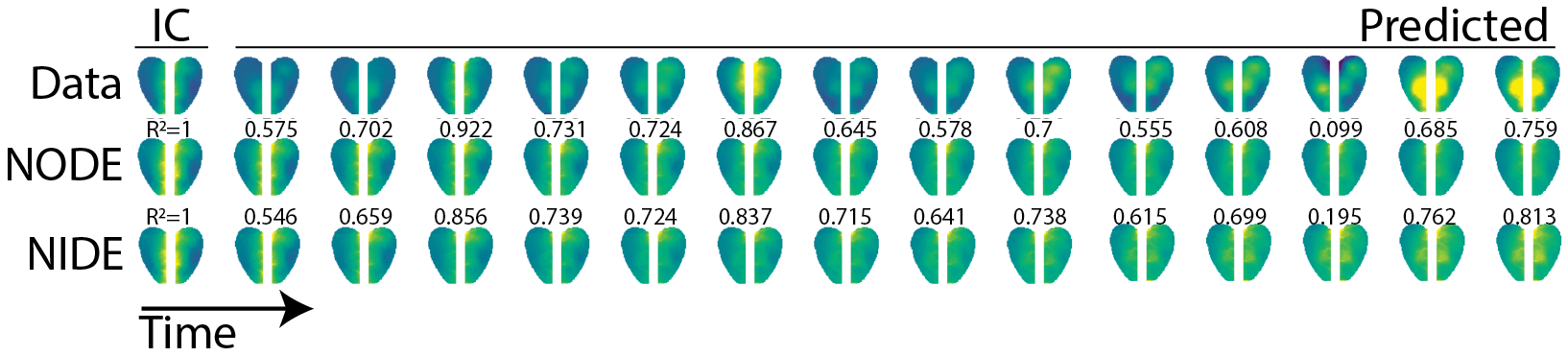}
	\end{center}
	\caption{Comparison of NIDE to NODE on the task of predicting dynamics of new initial conditions in wide-field calcium brain imaging data. Models were trained on multiple dynamics, each with a different initial condition. Models were then tested on their ability to predict dynamics from initial conditions not seen during training. Shown are dynamics for the original data (top), NODE prediction (middle), and NIDE prediction (bottom). $R^2$ values of predictions to ground truth are shown above each frame. NIDE outperforms NODE in this task. IC signifies the initial condition frame which is shown to the models.}
	\label{fig:brain_new_init}
\end{figure*}

\begin{table*}
	\caption{\ Specification of the models used for each experiment. Here, the networks F and K have the same architecture while the Free function consists of one hidden layer with 40 nodes.}
	\label{tab:Hyperparameters}
	\centering
	\begin{tabular}{lccccc}
		\toprule
		& F architecture & F & Kernel & Free function \\
		\midrule
		2D curves & [25,50,100,50,25] & 12954  & 13004 & - \\
		4D curves & [100,100,100,100,100] &  41304 & 41504 & 3684  \\
		Calcium Imaging & [100,100,100,100,100] &  42510 & 42710 & 4170 \\
		fMRI  & [100,100,100,100,100] &  42510 & 42710 & 4170   \\
	\end{tabular}
\end{table*}

\begin{table*}
	\caption{MSE per extrapolated time point on 4D curves dataset}.
	\label{tab:4d_curves_extrapolation}
	\centering
	\begin{tabular}{lccccccccccccccc}
		\toprule
		& \multicolumn{7}{c}{Extrapolated point }                   \\
		\cmidrule(r){2-8}
		& t+1 & t+2 & t+3 & t+4 & t+5 & t+6 & t+7 \\
		\midrule
		LSTM & 1.49E+00 & 1.53E+00 & 1.62E+00 & 1.64E+00 & 1.75E+00 & 1.98E+00 & 2.38E+00 \\ 
		NODE  & 1.57E-03 & 2.86E-03 & 5.91E-03 & 1.37E-02 & 3.59E-02 & 9.77E-02 & 2.55E-01 \\ 
		NIDE & \textbf{1.38E-03} & \textbf{2.58E-03} & \textbf{5.51E-03} & \textbf{1.28E-02} & \textbf{3.37E-02} & \textbf{9.23E-02} & \textbf{2.45E-01} \\ 
	\end{tabular}
\end{table*}

\begin{table*}
	\caption{\ Mean $R^2$ per extrapolated calcium imaging frame}
	\label{tab:brain_extrapolation}
	\centering
	\begin{tabular}{lccccc}
		\toprule
		& \multicolumn{3}{c}{Extrapolated frame }                   \\
		\cmidrule(r){2-3}
		& t+1 & t+2 \\
		\midrule
		LSTM & 0.6535 & 0.6376 \\
		NODE & 0.6577 & 0.6672 \\
		NIDE & \textbf{0.6767} & \textbf{0.6843}  \\
	\end{tabular}
\end{table*}

\begin{table*}
	\caption{MSE for extrapolated points of the dynamics for the $4D$ curves dataset. The initial points are used for training ($t+0$ to $t+4$) and are therefore seen by the models during training, while points $t+5$ to $t+11$ are extrapolated from the learned models.}
	\label{tab:4d_new_initial_conditions}
	\centering
	\begin{tabular}{lccccccccccccccc}
		\toprule
		& \multicolumn{7}{c}{Extrapolated point}                   \\
		\cmidrule(r){2-8}
		& t+5 & t+6 & t+7 & t+8 & t+9 & t+10 & t+11 \\
		\midrule
        LSTM & 4.42E-03 & 1.29E-02 & 3.04E-02 & 5.95-E02 & 1.13E-01 & 2.08E-01 & 3.80E-01  \\ 
		NODE  & 2.59E-03 & 6.09E-03 & 1.41E-02 & 3.05E-02 & 6.51E-02 & 1.36E-01 & 2.78E-01  \\ 
		NIDE &  \textbf{2.43E-03} & \textbf{5.42E-03} & \textbf{1.24E-02} & \textbf{2.68E-02} & \textbf{5.83E-02} & \textbf{1.25E-01} & \textbf{2.61E-01}  \\ 
	\end{tabular}
\end{table*}

\begin{figure*}[htb]
	\begin{center}
		\includegraphics[width=5in]{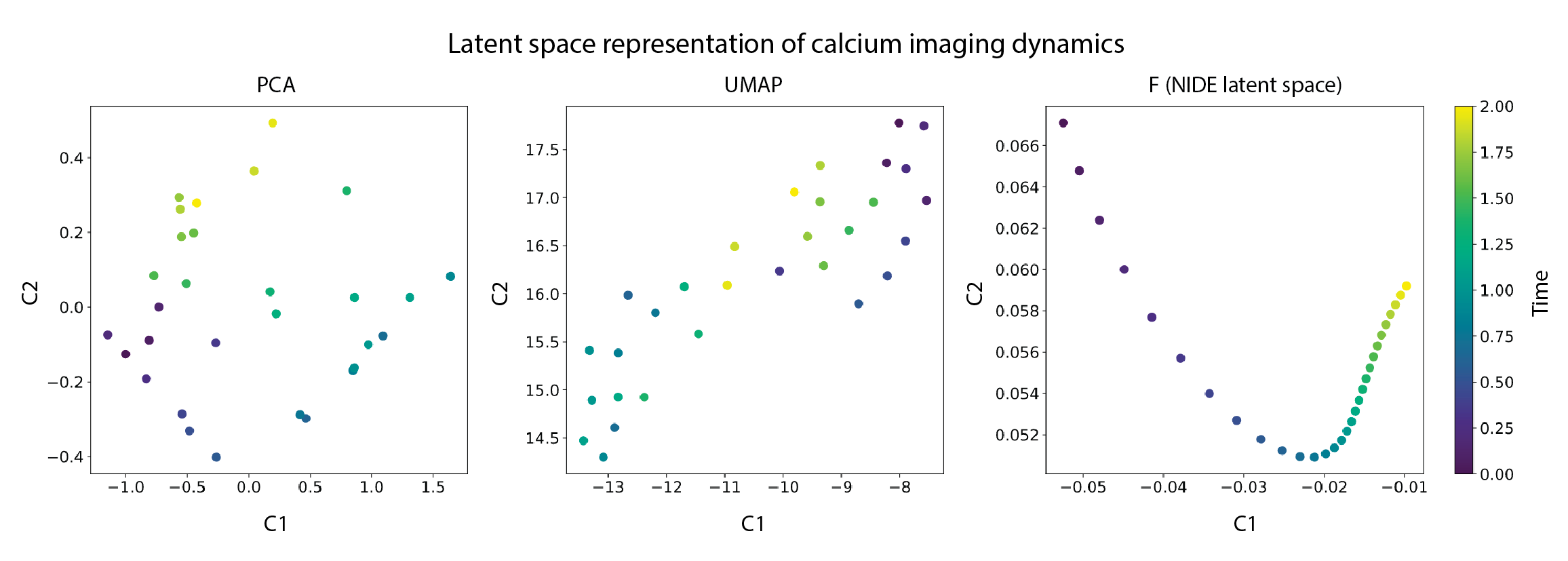}
	\end{center}
	\caption{Comparison of latent spaces of calcium brain imaging dynamics obtained via PCA, UMAP, and NIDE (via $F$ integrand). NIDE provides the most organized latent space, as quantified by k-NN regression (k=3 neighbors) against time ($R^2_{PCA}=0.88$, $R^2_{UMAP}=0.79$, $R^2_{NIDE}=0.99$)}
	\label{fig:brain_latent}
\end{figure*}

\begin{figure*}[htb]
	\begin{center}
		\includegraphics[width=5in]{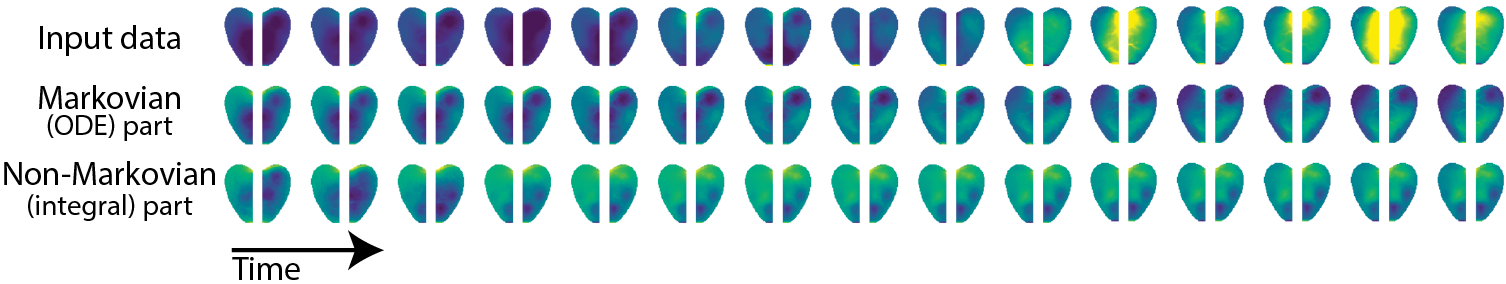}
	\end{center}
	\caption{Decomposition into Markovian and non-Markovian components with NIDE that was trained on wide-field calcium brain imaging dynamics of freely behaving mice. Top row shows the original data. Middle row shows the Markovian (ODE part) of the dynamics. Bottom row shows the non-Markovian (integral part) of the dynamics.}
	\label{fig:brain_markovian}
\end{figure*}

\begin{table*}
	\caption{Nearest neighbors accuracy on the classification of fMRI conditions in function of the number of neighbors (k)}
	\label{tab:knn_performances}
	\centering
	\begin{tabular}{lccccccccccccccc}
		\toprule
		& \multicolumn{6}{c}{Number of neighbors (k)}                   \\
		\cmidrule(r){2-5}
		& 2 & 3 & 4 & 5 \\
		\midrule
		Instantaneous  & \textbf{0.88} & \textbf{0.89} & \textbf{0.82} & \textbf{0.8}  \\ 
		Integral &  0.78 & 0.78 & 0.72 & 0.71  \\
		Raw &  0.66 & 0.73 & 0.57 & 0.60 \\
	\end{tabular}
\end{table*}

\section{Data collection}\label{sec:data_collection}

Here we report details about data collection and preprocessing. 

\subsection{Toy data}

The toy data has been obtained by solving analytical IDEs in $2D$ and $4D$s. The kernels used in both cases were convolutional kernels where the entries were given by combinations of trigonometric functions. The $F$ function was a hyperbolic cosine. To obtain the datasets, we have randomly sampled initial conditions and solved the corresponding initial value problem for the IDEs, using our implementation of the IDE solver. The integrals have been performed with Monte-Carlo integration with $1K$ sampled points per interval, and the number of iterations used was set to $10$, which was empirically seen to guarantee convergence to the solution.

\subsection{Calcium imaging dataset}
C57BL/6J mice were kept on a 12h light/dark cycle, provided with food and water ad libitum, and housed individually following headpost implants. Imaging experiments were performed during the light phase of the cycle. For mesoscopic imaging, brain-wide expression of jRCaMP1b was achieved via postnatal sinus injection as described in  \cite{barson2020simultaneous,hamodi2020transverse}. 

Briefly, P0-P1 litters were removed from their home cage and placed on a heating pad. Pups were kept on ice for 5 min to induce anesthesia via hypothermia and then maintained on a metal plate surrounded by ice for the duration of the injection. Pups were injected bilaterally with 4 ul of AAV9-hsyn-NES-jRCaMP1b ($2.5\times10^{13}$ gc/ml, Addgene). Mice also received an injection of AAV9-hsyn-ACh3.0 to express the genetically encoded cholinergic sensor ACh3.0, (Jing et al., 2020, although these data were not used in the present study. Once the entire litter was injected, pups were returned to their home cage.

Surgical procedures were performed on sinus injected animals once they reached adulthood ($>$P50). Mice were anesthetized using 1-2\% isoflurane and maintained at 37ºC for the duration of the surgery. For mesoscopic imaging, the skin and fascia above the skull were removed from the nasal bone to the posterior of the intraparietal bone and laterally between the temporal muscles. The surface of the skull was thoroughly cleaned with saline and the edges of the incision secured to the skull with Vetbond. A custom titanium headpost for head fixation was secured to the posterior of the nasal bone with transparent dental cement (Metabond, Parkell), and a thin layer of dental cement was applied to the entire dorsal surface of the skull. Next, a layer of cyanoacrylate (Maxi-Cure, Bob Smith Industries) was used to cover the skull and left to cure ~30 min at room temperature to provide a smooth surface for trans-cranial imaging.

Mesoscopic calcium imaging was performed using a Zeiss Axiozoom with a 1x, 0.25 NA objective with a 56 mm working distance (Zeiss). Epifluorescent excitation was provided by an LED bank (Spectra X Light Engine, Lumencor) using two output wavelengths: 395/25 (isosbestic for ACh3.0, Lohani et al., 2020) and 575/25nm (jRCaMP1b). Emitted light passed through a dual camera image splitter (TwinCam, Cairn Research) then through either a 525/50 (ACh3.0) or 630/75 (jRCaMP1b) emission filter (Chroma) before it reached two sCMOS cameras (Orca-Flash V3, Hamamatsu). Images were acquired at 512x512 resolution after 4x pixel binning. Each channel was acquired at 10 Hz with 20 ms exposure using HCImage software (Hamamatsu).

For visual stimulation, sinusoidal drifting gratings (2 Hz, 0.04 cycles/degree were generated using custom-written functions based on Psychtoolbox in Matlab and presented on an LCD monitor at a distance of 20 cm from the right eye. Stimuli were presented for 2 seconds with a 5 second inter-stimulus interval.

Imaging frames were grouped by excitation wavelength (395nm, 470nm, and 575nm) and downsampled from 512$\times$512 to 256$\times$256 pixels. Detrending was applied using a low pass filter (N=100, $f_{cutoff}=$0.001Hz). Time traces were obtained using $(\Delta F/F)_i=(F_i-F_{(i,o)} )/F_{(i,o)}$ where $F_i$ is the fluorescence of pixel $i$ and $F_{(i,o)}$ is the corresponding low-pass filtered signal.

Hemodynamic artifacts were removed using a linear regression accounting for spatiotemporal dependencies between neighboring pixels.  We used the isosbestic excitation of ACh3.0 (395 nm) co-expressed in these mice as a means of measuring activity-independent fluctuations in fluorescence associated with hemodynamic signals.  Briefly, given two $p\times1$  random signals $y_1$ and $y_2$ corresponding to $\Delta F/F$ of $p$ pixels for two excitation wavelengths “green” and "UV", we consider the following linear model:

\begin{eqnarray}
y_1=x+z+\eta,
\end{eqnarray}
\begin{eqnarray}
y_2=Az+\xi,
\end{eqnarray}	

where x and z are mutually uncorrelated $p\times1$ random signals corresponding to $p$ pixels of the neuronal and hemodynamic signals, respectively. $\eta$ and $\xi$ are white Gaussian $p\times1$ noise signals and A is an unknown $p\times p$ real invertible matrix. We estimate the neuronal signal as the optimal linear estimator for $x$ (in the sense of Minimum Mean Squared Error):

\begin{eqnarray}
\hat{x} = H \begin{pmatrix} y_1  \\ y_2  \end{pmatrix}, 
H = \sum_{xy}{\sum_{y}}^{-1}
\end{eqnarray}

where $y=\begin{pmatrix} y_1  \\ y_2  \end{pmatrix}$ is given by stacking $y_1$  on top of $y_2$,  $\sum_y=E[yy^T ]$ is the autocorrelation matrix of $y$ and $\sum_{xy}=E[xy^T ]$ is the cross-correlation matrix between $x$ and $y$. The matrix $\sum_y$ is estimated directly from the observations, and the matrix $\sum_{xy}$ is estimated by:

\begin{eqnarray}
\sum_{xy}=\Biggl(\sum_{y_1}
-\sigma_{\eta}^{2}I- \biggl(\sum_{y_1 y_2} {\Bigl(\sum_{y_2}-\sigma_{\xi}^{2}I \Bigl)}^{-1}{\sum_{y_2}}^{-1} {\sum_{y_1 y_2}}^T \biggl)^T\Bigg) \notag
\end{eqnarray}

where $\sigma_\eta^2$ and $\sigma_\xi^2$  are the noise variances of $\eta$ and $\xi$, respectively, and $I$ is the $p\times p$ identity matrix. The noise variances $\sigma_\eta^2$ and $\sigma_\xi^2$  are evaluated according to the median of the singular values of the corresponding correlation matrices  $\sum_{y_1}$and $\sum_{y_2}$.  This analysis is usually performed in patches where the size of the patch, $p$, is determined by the amount of time samples available and estimated parameters. In the present study, we used a patch size of $p=9$.   The final activity traces were obtained by z-scoring the corrected $\Delta F/F$ signals per pixel. The dimensionality of the resulting video is then reduced via PCA to 10 components, which represents $\approx 80\%$ of data variance.

\subsection{Functional magnetic resonance imaging (fMRI) dataset}
Saline study structural (1x1x1 $mm^3$) and functional (3.2x3.2x3.1 $mm^3$; TR=2000 $ms$; TE=28 $ms$; 12 min. at baseline and week 1) scans, and ketamine study structural (1x1x1 $mm^3$) and functional (3x3x2.5 $mm^3$ ; TR=3000 ms.; TE=30 ms.; 5min. immediately prior to infusion and 20 min. during infusion starting at 20 min post administration) scans were acquired using 3.0 T magnets \cite{abdallah2018ketamine}. Brain scans from both studies underwent the same surface-based preprocessing using pipeline adapted from the Human Connectome Project \cite{glasser2013minimal}. Briefly, the preprocessing pipeline included FreeSurfer parcellation of structural scans, slice timing correction, motion correction, intensity normalization, brain masking, and registration of fMRI images to structural MRI and standard template. Then, the cortical gray matter ribbon voxels and each subcortical parcel were projected to a standard Connectivity Informatics Technology Initiative (CIFTI) 2mm grayordinate space. ICA-FIX was run to identify and remove artifacts, followed by mean grayordinate time series regression (MGTR) \cite{griffanti2014ica, burgess2016evaluation}. 

The brain nodes were defined using  multimodal parcellation atlases that divide the cerebral cortex, subcortical regions, and the cerebellum into 424 nodes within the grayordinate. Within each node, an averaged time series of all voxels/vertices was calculated. The atlases used for cortex, subcortical regions and cerebellum parcellation are described in \cite{glasser2016multi, fan2016human, diedrichsen2009probabilistic}.

\end{document}